\pdfoutput=1
\documentclass[]{ceurart}

\usepackage{listings}
\usepackage{booktabs}
\usepackage{caption}
\usepackage{subcaption}
\begin{document}

\copyrightyear{2026}
\copyrightclause{Copyright for this paper by its authors.
  Use permitted under Creative Commons License Attribution 4.0
  International (CC BY 4.0).}

\conference{CLEF 2026 Working Notes, 21 -- 24 September 2026, Jena, Germany}

\title{Overview of the ImageCLEFmedical 2026 GANs task: Exploring fingerprint detection, privacy leakage, and synthetic medical image generation}
\title[mode=sub]{ImageCLEF at CLEF 2026}

\author[1]{Alexandra{-}Georgiana Andrei}[email=alexandra.andrei@upb.ro]
\cormark[1]
\author[1]{Mihai Gabriel Constantin}[email=mihai.constantin84@upb.ro]
\author[1]{Mihai Dogariu}[email=mihai.dogariu@upb.ro]
\author[2]{Dzmitry Karpenka}[email=karpenko.dima.s11@gmail.com]
\author[2]{Yuri Prokopchuk}[email=prokopchuk@newman.bas-net.by]
\author[2]{Ahmedkhan Radzhabov}[email=axmegxah@outlook.com]
\author[1]{Dan-Cristian Stanciu}[email=dan.stanciu1203@upb.ro]
\author[1]{Liviu-Daniel \textcommabelow{S}tefan}[email=liviu_daniel.stefan@upb.ro]
\author[2]{Vassili Kovalev}[email=vassili.kovalev@gmail.com]
\author[3]{Henning M\"uller}[email=henning.mueller@hevs.ch]
\author[1]{Bogdan Ionescu}[email=bogdan.ionescu@upb.ro]

\address[1]{AI Multimedia Lab, CAMPUS Research Institute, National University of Science and Technology Politehnica Bucharest, Romania}
\address[2]{National Academy of Science of Belarus, Minsk, Belarus}
\address[3]{University of Applied Sciences Western Switzerland (HES-SO), Sierre, Switzerland}

\cortext[1]{Corresponding author.}

\begin{abstract}
The 2026 ImageCLEFmedical GANs Task Exploring Fingerprint Detection, Privacy Leakage, and Synthetic Medical Image Generation continues to investigate privacy and security concerns around using patient data to generate synthetic medical images, now in its fourth edition. This edition introduces three complementary subtasks: the first extends prior editions by asking participants to detect which real images were used in training a Generative Adversarial Network to produce given synthetic outputs; the second is a novel subtask requiring participants to identify which latent vector produced each synthetic image, probing the faithfulness and reversibility of the latent-to-image mapping; the third asks participants to generate privacy-preserving CT slices, quantifying the trade-off between image fidelity and patient privacy. Ground-truth annotations and benchmark datasets of real and GAN-generated lung CT slices are provided for all subtasks. Evaluation is based on Cohen's Kappa for Subtask 1, matching accuracy for Subtask 2, and a combination of FID and Privacy Preservation Score (PPS) for Subtask 3. Overall, 37 teams registered for the task, with 10 teams submitting runs across subtasks and 6 teams submitting working notes papers. This paper presents an overview of the task setup, datasets, evaluation metrics, and summarises the approaches and results of participating teams.
\end{abstract}


\maketitle

\section{Introduction}
\label{sec:introduction}

AI and computer-aided diagnosis systems for medical tasks such as prediction, detection, and classification depend on access to large and diverse datasets for effective training. High-quality data enable these models to learn complex patterns, improving both accuracy and reliability. However, obtaining real medical data is difficult due to strict privacy regulations and ethical concerns. Patients typically consent to the use of their medical information for their own clinical care, but not for broad research purposes. As a result, collecting sufficiently large datasets for training AI models remains a significant challenge and slows progress in developing advanced healthcare tools.

Synthetic medical data has emerged as a promising solution~\cite{frid2018gan}. Generative models such as Generative Adversarial Networks (GANs)~\cite{goodfellow2014generative} can produce realistic-looking images that resemble true medical data without being directly tied to any specific patient, potentially easing data-access barriers and enabling greater diversity and scalability. However, the privacy guarantees of synthetic data are not unconditional. Generative models are trained on real patient images, and may inadvertently retain hidden traces or ``fingerprints'' of their training data in the images they produce. If synthetic images can be linked back to specific patients 
or reveal patient-specific anatomical details, the privacy benefit of using synthetic data is fundamentally undermined. Ensuring that generative models do not memorise or propagate identifiable patterns from their training data is therefore a critical open problem, and one that must be rigorously studied before synthetic medical images can be safely deployed in research and clinical AI development.

This is the fourth edition of the GANs task, part of ImageCLEF 2026~\cite{ImageCLEF2026,ionescu2026imageclef}, in which we continue to investigate the hypothesis that generative models generate synthetic medical images that retain ``fingerprints'' from the real images used during their training, based on lessons learned from the previous three editions:

\begin{itemize}
  \item In the first~\cite{GAN2023} and second~\cite{GAN2024} editions of this task, held at ImageCLEF 2023~\cite{ImageCLEF2023} and 2024~\cite{ImageCLEF2024}, various generative models were analysed within the framework of the first subtask to investigate whether synthetic images contained ``fingerprints'' of the real medical data used during training. The results demonstrated that the tested generative models do retain and imprint features from their training data, raising important security and privacy concerns.
  \item In the 2nd edition of the task~\cite{GAN2024}, it was confirmed that generative models leave unique ``fingerprints'' on the synthetic images they produce. By analysing images generated from various models, distinct patterns and features were identified that allowed the attribution of synthetic images to their respective generative models.
  \item In the 3rd edition~\cite{GAN2025} held at ImageCLEF 2025~\cite{ImageCLEF2025}, the subtask ``Identify Training Data Subsets'' explored attribution at a broader scale: identifying the training subset used to generate synthetic images from a Diffusion model. Performance was significantly higher, with multiple teams exceeding 98\% accuracy, revealing that diffusion-generated images can still reflect statistical characteristics of the training data subsets and may encode latent information about their source data distributions.
\end{itemize}

To investigate the potential privacy risks associated with synthetic medical images, the 2026 edition of the GANs Task focuses on three complementary subtasks. The first subtask challenges participants to determine whether specific real images were used during the training of a GAN to produce given synthetic outputs, addressing the risk of image-level information leakage. The second subtask is entirely new: participants must match each synthetic image to its originating 128-dimensional latent vector, probing whether the latent-to-image mapping is reversible and whether latent space encodes patient-specific information. The third subtask invites participants to train generative models and submit synthetic CT slices, which are evaluated for both image fidelity and privacy preservation using four independent privacy attacks.

\section{Task description}
\label{sec:taskdescription}

\subsection{Subtask 1: Detect training data usage}

\subsubsection{Description}
We continue to investigate the hypothesis that generative models generate medical images that are in some way similar to the ones used for training. The task addresses the security and privacy concerns related to personal medical image data in the context of generating and using artificial images in different real-life scenarios. This edition continues the task proposed in all previous editions~\cite{GAN2023, GAN2024, GAN2025} by investigating another GAN. The objective is to detect ``fingerprints'' within the synthetic biomedical image data to determine which real images were used in training to produce the generated images. The task is formulated as follows:

\begin{itemize}
  \item \textit{In this subtask, participants will analyse synthetic biomedical images to determine whether specific real images were used in the training process of generative models. For each real image in the test set, participants must assign a binary label: 1 if the image was used during training or 0 if the image was not used. The goal is to detect the presence of training-data ``fingerprints'' within synthetic outputs and to assess vulnerabilities related to memorisation.}
\end{itemize}

\subsubsection{Data description}
The benchmarking dataset includes both real and synthetic biomedical images. The real images consist of axial slices of 3D thoracic CT scans from approximately 8,000 lung tuberculosis patients. These slices vary in appearance: some may look relatively ``normal'', while others exhibit distinct lung lesions, including severe cases. The real images were provided in 8-bit per pixel PNG format, with dimensions of 256$\times$256 pixels.

The development dataset consists of 10,000 synthetic images generated using the studied GAN and 6,400 annotated real images (3,200 used for training (label 1); 3,200 not used for training (label 0)).

The test dataset contains 2,140 real CT images with unknown labels, including a mix of images that were and were not used to train the GAN.

\subsubsection{Evaluation protocol}
The primary evaluation metric is Cohen's Kappa, which measures the agreement between predicted and ground-truth labels while accounting for chance agreement, making it robust to class imbalance. In addition, the following metrics are reported: Accuracy, Sensitivity (Recall/TPR), Specificity (TNR), Precision, F1 Score, and Matthews Correlation Coefficient (MCC). All metrics are computed with label 1 (used) as the positive class. Cohen's kappa coefficient ranges from -1 (complete disagreement) to 1 (complete agreement)~\cite{mchugh2012interrater}, and is defined as:

\begin{equation}
  k = (p_o - p_e) / (1 - p_e)
\end{equation}

where $p_o$ is the empirical probability of agreement on the label assigned to any sample (the observed agreement ratio), and $p_e$ is estimated using a per-annotator empirical prior over the class labels.

\subsection{Subtask 2: Identify corresponding latent vectors}

\subsubsection{Description}
This subtask evaluates how faithfully generative models map latent vectors to output images, and whether this mapping reveals structural or patient-specific information. Participants are given a set of synthetic images alongside a collection of latent vectors and must establish the correspondence between each generated image and its originating latent vector. The latent vectors are 128-dimensional and sampled from a standard normal distribution $\mathcal{N}(0,1)$.

The task is formulated as follows:
\begin{itemize}
  \item \textit{In this subtask, participants will analyse a set of synthetic images and a collection of latent vectors (128-dimensional, sampled from a standard normal distribution) and determine which latent vector corresponds to which generated image. This task evaluates how faithfully generative models map latent vectors to output images and whether this mapping reveals structural or patient-specific information.}
\end{itemize}

A well-generalising model should produce synthetic images that do not reflect identifiable traces of specific training samples, even when latent vectors are disclosed. In contrast, an overfitted model may encode training-data fingerprints within the latent space, enabling reconstruction or linkage back to individuals.
By requiring participants to match images to latent vectors, this subtask measures: (i) the predictability of the latent-to-image mapping; (ii) the presence of hidden structure or memorisation within the latent space; and (iii) the robustness of generative models to latent-space-based privacy attacks. This subtask therefore complements Subtask 1 by examining privacy leakage not only through image similarity, but through learned representations.

\subsubsection{Data description}
The development dataset consists of 10,000 image--latent vector pairs generated by the BigGAN-Deep model~\cite{brock2018large} at 256$\times$256 resolution. Each latent vector $z \in \mathbb{R}^{128}$ is independently sampled from a standard normal distribution $\mathcal{N}(0, I_{128})$. 

The test set consists of a set of GAN-generated images and a file containing the latent vectors. The goal is to identify, for each latent vector, the image generated from it. 

\subsubsection{Evaluation protocol}
The primary evaluation metric is Matching Accuracy: the proportion of submitted image--latent vector pairs for which the predicted index exactly matches the ground truth.

\subsection{Subtask 3: Privacy-preserving CT slice generation}

\subsubsection{Description}
Generative models trained on medical data may unintentionally memorise and reproduce sensitive patient information. In this subtask, participants are asked to generate new CT slices using a set of real slices as guidance. Although the task resembles a typical image-generation problem, its primary goal is to study privacy leakage and fingerprint propagation by investigating two complementary hypotheses: (i) models that overfit may reconstruct patient-specific anatomy or replicate identifiable features; (ii) models that generalise well should produce realistic CT images without encoding individual patient identities.
The objective of Subtask 3 is to develop generative models capable of producing realistic and diverse CT slices while preserving patient privacy. The task focuses on learning general anatomical structure rather than reproducing or memorising specific training images. This subtask allows us to quantify the trade-off between realism and privacy preservation.

\subsubsection{Data description}
The dataset for this subtask consists of 10,000 256$\times$256 two-dimensional axial lung CT slices provided for model training. The development dataset contains real CT slices intended for model training; all images are 2D axial slices drawn from multiple patients from the same cohort described in Subtask 1, with no patient identifiers provided. Participants submit 5,000 generated synthetic CT slices together with their generator checkpoint.

\subsubsection{Evaluation protocol}
Submitted images are assessed across two dimensions: quality and privacy.
Image quality is measured using the Fréchet Inception Distance (FID), computed with a task-specific feature extractor trained on domain-relevant images.

Privacy is quantified through a Privacy Preservation Score (PPS), defined as a composite score averaging four leakage scores derived from four distinct attack types. All attacks use a held-out patient set, CT slices from patients never released to participants, as a negative control.

\noindent\textbf{(i) Attack 1 - Nearest-Neighbour Distance (NND)} ($\ell_1$): measures instance-level memorisation, i.e. whether any synthetic images are near-copies of specific training samples. For each synthetic image, the distance to its nearest neighbour is computed separately in the training set and in a held-out set, using deep InceptionV3 feature representations. The key metric is the NND ratio:
\[
    R = \frac {mean\;distance\;(synthetic \rightarrow training)}{mean\; distance\;(synthetic \rightarrow held\text{-}out)},\;\; \ell_1=clip\;(1-R, 0, 1)
\]

Images whose distance to the training set falls below a memorisation threshold are individually flagged.

\noindent\textbf{(ii) Attack 2 - Membership Inference Attack (MIA)} ($\ell_2$): measures distributional bias, i.e. whether the synthetic distribution as a whole reveals which patients were used for training. Each real CT image is labelled as a member (from the training set) or non-member (from the held-out set). A membership score is assigned based on how close each real image is to its nearest synthetic neighbour in feature space. The AUC-ROC of this binary classifier measures how reliably an adversary can distinguish training patients from held-out patients using only the submitted synthetic images.
TPR@5\%FPR is additionally reported: the fraction of training images correctly identified at a 5\% false-positive rate.
\[
 \ell_2=clip\;((AUC-0.5)\;/\;0.5,\; 0,\; 1)
\]
\textbf{(iii) Attack 3 - Patient Re-Identification} ($\ell_3$): measures patient identity leakage, i.e. whether the synthetic distribution reflects the anatomical characteristics of individual training patients, even if no image is a direct visual copy. A per-patient anatomical centroid is built by averaging the feature vectors of all training slices from each patient. Each synthetic image is assigned to the most similar patient centroid via cosine similarity. The key metric is rank-1 patient coverage: the fraction of training patients that appear as the top match for at least one synthetic image.
\[
 \ell_3=clip\;(rank\text{-}1\;coverage,\; 0,\; 1)
\]
\textbf{(iv) Attack 4 - Generalised Inversion Attack} ($\ell_4$): measures latent-space memorisation, i.e. whether the generator's internal representations encode training-specific anatomy in a way that allows training images to be reconstructed more faithfully than unseen held-out images. Using the submitted generator checkpoint, gradient-based optimisation finds the input noise that best reconstructs each target real image. This is applied separately to a sample of training images and held-out images. The key metric is the reconstruction delta:
\[
\Delta= mean\;reconstruction\;error\;(held\text{-}out) - mean\;reconstruction\;error\;(training)
\]
\[
 \ell_4=clip\;(\Delta\;/\;mean\;reconstruction\;error\;(held\text{-}out)\;/\;0.5,\; 0,\; 1)
\]

The PPS is computed as:
\begin{equation}
  \text{PPS} = 1 - \text{mean}(\ell_1, \ell_2, \ell_3, \ell_4)
\end{equation}

A higher PPS indicates stronger privacy preservation.

\section{Results}
\label{sec:results}

\subsection{Participation statistics}

Overall, 37 teams registered for the 2026 edition of the task, with 6 teams submitting working notes papers. Table~\ref{tab:teams} presents the list of participating teams that submitted working notes papers and their affiliations. Compared to the previous edition~\cite{GAN2025}, which attracted 41 registered teams with 14 teams completing Subtask 1, 4 teams completing Subtask 2, and 9 teams submitting working notes papers, participation in the 2026 edition reflects the increased complexity introduced by the two new subtasks. The continued participation of teams from diverse geographic regions reflects the sustained international interest in the privacy and security challenges associated with synthetic medical image generation.

\begin{table}[!tp]
\caption{Overview of participating teams that submitted working notes papers (* task organising team).}
\label{tab:teams}
\centering
\begin{tabular}{lcccp{4cm}l}
\toprule
Team & Subtask~1 & Subtask~2 & Subtask~3 & Affiliation & Country \\
\midrule
Bharathi \& Harish~\cite{harishandbharathi}& \checkmark & & & Shiv Nadar University & India \\
csmorgan~\cite{csmorgan} & \checkmark & \checkmark & \checkmark & Morgan State University & USA \\
CVG-IBA~\cite{CVG-IBA} & \checkmark & \checkmark & & Institute of Business Administration, Karachi & Pakistan \\
DS@GT ARC~\cite{eric} & & & \checkmark & Georgia Institute of Technology & USA \\
Agentili~\cite{agentili} & & \checkmark & & San Diego VA Health Care Center / UC San Diego & USA \\
AI Multimedia Lab*~\cite{AIM} & \checkmark & \checkmark & \checkmark & National University of Science and Technology POLITEHNICA Bucharest & Romania \\
\bottomrule
\end{tabular}
\end{table}

\subsection{Subtask 1: Detect training data usage}

\subsubsection{Presented methods}

\textbf{Bharathi \& Harish}~\cite{harishandbharathi} investigated transfer learning approaches using pre-trained deep learning models. The team formulated the task as a supervised binary classification problem, training only on the 6,400 labelled real images from \texttt{real\_used} and \texttt{real\_not\_used}, without using the generated images. Three backbone models were fine-tuned: ResNet34, with layers layer3 and layer4 unfrozen and a two-class output head; EfficientNet-B3, with the last two feature blocks unfrozen and a Dropout-augmented classifier; and DenseNet121, with denseblock4 unfrozen. All models used Adam optimisation (lr=0.0003), CrossEntropyLoss with class weights [2.0, 1.0], StepLR scheduling, and 10 training epochs. An ensemble of the three models using soft voting (averaging softmax probabilities) was submitted as the final run.
\\

\noindent\textbf{Csmorgan}~\cite{csmorgan} adopted a stacked feature-based ensemble combining four complementary information sources. Azimuthally averaged log power spectra of real images were compared to the statistics of synthetic images, yielding spectral distance and $z$-score features. Low-level image descriptors including intensity statistics (mean, std, skewness, kurtosis, percentiles), histogram entropy, and Sobel edge magnitude were also computed directly from the normalised grayscale pixel values. High-level deep features were extracted from the conv-3, conv-4, and conv-5 stages of an ImageNet-pretrained ResNet-50, computing cosine and $\ell_2$ distances to the $k=1,3,5,10,20$ nearest neighbours in the synthetic pool. Shadow-autoencoder reconstruction error features were also included. The final feature vector was passed to an ensemble of HistGradientBoosting classifiers, a GradientBoostingClassifier, and an ExtraTreesClassifier, trained under 5-fold stratified cross-validation. The decision threshold was selected by maximising Cohen's $\kappa$ on out-of-fold predictions.
\\

\noindent\textbf{CVG-IBA}~\cite{CVG-IBA} evaluated three independent detectors combined in four ensemble configurations. M1 was a fine-tuned ResNet-18 classifier with a custom head (FC $512\rightarrow256\rightarrow2$), trained with AdamW, One-Cycle LR, and augmentations including horizontal/vertical flip, affine transforms, colour jitter, and Gaussian blur. M2 was a handcrafted pixel-statistics and GAN-fingerprint Random Forest, with features comprising intensity histograms, global statistics, LBP, GLCM Haralick features, gradient statistics, DCT energy, FFT radial energy, and a GAN fingerprint correlation coefficient. M3 was a bagged ensemble of $K=5$ Convolutional Autoencoders trained on GAN-generated images, producing a membership score from reconstruction error statistics. Ensemble configurations E1--E4 combined these detectors by averaging class probability vectors.
\\

\noindent\textbf{AI Multimedia Lab}~\cite{AIM} proposed a membership inference pipeline with four stages: deep feature extraction using a pretrained ResNet-50 fixed encoder (2048-dimensional embeddings); $k$-nearest-neighbour distance computation in the generated image feature space ($k=5$), deriving four distance statistics; pixel-level descriptor computation (mean intensity, contrast, skewness proxy, entropy proxy); and classification using a Gradient Boosting classifier (200 estimators, depth 4, lr=0.05) under 5-fold stratified cross-validation. A simple threshold-based rule applied directly to the mean $k$-NN distance was also evaluated as an interpretable baseline.

\subsubsection{Results}
Table~\ref{tab:subtask1} reports the official test-phase results for all Subtask 1 submissions with working notes.
Cohen's $\kappa$ scores across all submissions were negative, ranging from $-0.0804$ (csmorgan, Runs 465 and 469) to $-0.0243$ (CVG-IBA, run 1170). The best-ranked submission among participating teams teams was CVG-IBA (Run 1170, $\kappa=-0.0243$, accuracy=0.4879), followed by csmorgan Run 467 ($\kappa=-0.0290$, accuracy=0.4855) and Bharathi \& Harish Run 843 ($\kappa=-0.0383$, accuracy=0.4808). Accuracy values across all runs ranged narrowly between 0.4598 and 0.4972, all below the 0.5 chance baseline. A notable pattern across submissions is the trade-off between sensitivity and specificity: csmorgan runs 467 and 466 achieved high specificity (0.9262 and 0.7075 respectively) at the cost of very low sensitivity (0.0449 and 0.2299), while Bharathi \& Harish runs 843 and 857 showed more balanced sensitivity and specificity values (0.4290/0.5327 and 0.4150/0.5075). The AI Multimedia Lab submission achieved the most balanced result overall, with sensitivity of 0.4935 and specificity of 0.5009, yielding the highest accuracy of 0.4972 and F1 score of 0.4953 across all submissions.

\begin{table}[!tp]
\caption{Results of participant submissions for Subtask 1: Detect Training Data Usage. (* task organising team)}
\label{tab:subtask1}
\centering
\begin{tabular}{llrrrrrrr}
\toprule
\# & Participant & Run ID & Cohen's $\kappa$ & MCC & Acc. & Sensitivity & Specificity & F1 \\
\midrule

1 & CVG-IBA & 1170 & -0.0243 & -0.0380 & 0.4879 & 0.1037 & 0.8720 & 0.1684 \\
2 & csmorgan & 467 & -0.0290 & -0.0613 & 0.4855 & 0.0449 & 0.9262 & 0.0802 \\
3 & Bharathi \& Harish & 843 & -0.0383 & -0.0385 & 0.4808 & 0.4290 & 0.5327 & 0.4524 \\
4 & csmorgan & 468 & -0.0570 & -0.0612 & 0.4715 & 0.2897 & 0.6533 & 0.3541 \\
5 & csmorgan & 466 & -0.0626 & -0.0713 & 0.4687 & 0.2299 & 0.7075 & 0.3020 \\
6 & Bharathi \& Harish & 857 & -0.0776 & -0.0779 & 0.4612 & 0.4150 & 0.5075 & 0.4351 \\
7 & csmorgan & 465 & -0.0804 & -0.0916 & 0.4598 & 0.2196 & 0.7000 & 0.2891 \\
8 & csmorgan & 469 & -0.0804 & -0.0916 & 0.4598 & 0.2196 & 0.7000 & 0.2891 \\
* & AI Multimedia Lab & -- & -0.0056 & -0.0056 & 0.4972 & 0.4935 & 0.5009 & 0.4953 \\
\bottomrule
\end{tabular}
\end{table}

\subsection{Subtask 2: Identify corresponding latent vectors}

\subsubsection{Presented methods}

\noindent\textbf{csmorgan}~\cite{csmorgan} framed Subtask 2 as a 10,000-way matching problem. The image branch used a ConvNeXt-Base backbone pretrained on ImageNet-22K, with a two-layer MLP projecting image features into a 256-dimensional embedding space. The latent branch used a three-layer MLP mapping each $z \in \mathbb{R}^{128}$ into the same space. Both branches produced $\ell_2$-normalised embeddings. Training used a memory-bank InfoNCE objective with temperature $\tau=0.05$ and MoCo-style momentum updates ($\rho=0.999$). At inference, optional post-processing included Sinkhorn normalisation, CSLS hubness correction, mutual-$k$NN reweighting, and multi-crop test-time augmentation; the final one-to-one assignment used the Hungarian algorithm. At inference, optional post-processing steps were applied: Sinkhorn normalisation to doubly-stochastically normalise the similarity matrix, CSLS (Cross-domain Similarity Local Scaling) hubness correction, mutual-$k$NN reweighting, and multi-crop test-time augmentation. Final one-to-one assignment used the Hungarian algorithm.
\\

\noindent\textbf{CVG-IBA}~\cite{CVG-IBA} explored two complementary paradigms: regression-based approaches that directly minimise distance between predicted and ground-truth $z$ vectors, and contrastive dual-encoder approaches that learn aligned image and $z$ embeddings. Three ResNet-50 regression variants were trained using different loss functions (MSE, cosine distance, and a convex combination), alongside two contrastive dual-encoder architectures (A4 with symmetric NT-Xent loss and batch size 256; A5 with partial backbone freeze, fixed temperature $\tau=0.2$, and extended training up to 100 epochs). All models were combined in five ensemble configurations by averaging pairwise similarity matrices before Hungarian assignment.
\\

\noindent\textbf{Agentili}~\cite{agentili} formulated the task as supervised image-to-latent regression: a deep convolutional network was trained to map each synthetic CT image to its corresponding 128-dimensional latent vector, after which predicted vectors were matched to the pool of candidates using cosine similarity and the Hungarian algorithm to enforce the one-to-one constraint. The development dataset of 10,000 BigGAN-Deep image--latent pairs was first split 80/20 for initial model development, then all 10,000 samples were used in 5-fold cross-validation for the final ensemble (8,000 training and 2,000 validation per fold). All images were loaded as RGB (grayscale replicated across three channels) and normalised using ImageNet statistics. Data augmentation during training comprised random horizontal flip ($p=0.5$) to exploit approximate left-right symmetry in axial CT views, and random rotation ($\pm15^\circ$) in selected model variants. Validation and test images were processed without augmentation. Several backbone architectures were evaluated (ResNet18, ResNet50, EfficientNetV2-S, ViT-B/16); EfficientNetV2-S proved consistently best. The training objective combined Huber loss and cosine embedding loss ($\mathcal{L} = \alpha \cdot \mathcal{L}_\text{Huber} + \mathcal{L}_\text{cos}$, with $\alpha \approx 0.10$--$0.15$), with the cosine term found critical for retrieval performance. Hyperparameters were tuned with Optuna (50 trials, TPE sampler). Five EfficientNetV2-S models were trained under 5-fold cross-validation, and predictions were averaged element-wise. Final assignment used the Hungarian algorithm on the $N \times N$ cosine similarity matrix, enforcing the one-to-one constraint.
\\

\noindent\textbf{AI Multimedia Lab}~\cite{AIM} explored four strategies. Method 1 (supervised encoder-based GAN inversion) fine-tuned a ResNet-50 backbone to directly predict latent vectors using a combined MSE and cosine loss, followed by Hungarian assignment on the cosine similarity matrix. Method 2 (dual encoder with symmetric InfoNCE) trained two separate encoders for images and $z$-vectors into a shared 256-dimensional hypersphere. Method 3 (improved dual encoder with hard negative mining, $z$-augmentation, and curriculum hybrid loss) extended Method 2 with a full-dataset memory bank and a regression weight decaying from 1.0 to 0.05 over 40 epochs. Method 4 (linear mapping in pretrained DINO feature space) fitted ridge regression models in both $z \rightarrow$ DINO and DINO $\rightarrow z$ directions, with PCA whitening on $z$-vectors and no training loop.
\begin{table}[!bp]
\caption{Results of participant submissions for Subtask 2: Identify Corresponding Latent Vectors.(* task organising team)}
\label{tab:subtask2}
\centering
\begin{tabular}{llrrrr}
\toprule
\# & Participant & Run ID & Accuracy & Correct & Wrong \\
\midrule
1 & Agentili & 1008 & 1.0000 & 10000 & 0 \\
2 & Agentili & 1009 & 0.9996 & 9996 & 4 \\
3 & Agentili & 909 & 0.9916 & 9916 & 84 \\
4 & CVG-IBA & 1169 & 0.3677 & 3677 & 6323 \\
5 & CVG-IBA & 1179 & 0.2135 & 2135 & 7865 \\
6 & CVG-IBA & 1178 & 0.1709 & 1709 & 8291 \\
7 & csmorgan & 591 & 0.0224 & 224 & 9776 \\
8 & csmorgan & 593 & 0.0108 & 108 & 9892 \\
9 & csmorgan & 592 & 0.0107 & 107 & 9893 \\
10 & csmorgan & 580 & 0.0094 & 94 & 9906 \\
11 & csmorgan	&	583&	0.0094&	94&	9906\\
12 &csmorgan	&582&	0.0048&	48& 9952\\
13 & csmorgan & 581 & 0.0003 & 3 & 9997 \\
* & AI Multimedia Lab & 965 & 0.3737 & 3737 & 6263 \\
* & AI Multimedia Lab & 968 & 0.0666 & 666 & 9334 \\
* & AI Multimedia Lab & 970 & 0.0105 & 105 & 9895 \\
* & AI Multimedia Lab & 969 & 0.0065 & 65 & 9935 \\

\bottomrule
\end{tabular}
\end{table}

\subsubsection{Results}

Table~\ref{tab:subtask2} reports the official test-phase results for Subtask 2. Matching accuracy across all Subtask 2 submissions spans a wide range, from 0.0003 (3 correct matches out of 10,000, csmorgan Run 581) to a perfect 1.0000 (10,000 correct matches, Agentili Run 1008). Agentili achieved the top three results in the ranking, with Runs 1008, 1009, and 909 obtaining matching accuracies of 1.0000, 0.9996, and 0.9916 respectively, correctly identifying all or nearly all image-latent correspondences. The next tier of performance was achieved by CVG-IBA and AI Multimedia Lab: CVG-IBA's best run (Run 1169) correctly matched 3,677 images (accuracy 0.3677), while the AI Multimedia Lab's best run (Run 965) correctly matched 3,737 images (accuracy 0.3737), with both teams' remaining submissions dropping substantially, to 0.2135 and 0.1709 for CVG-IBA and to 0.0666, 0.0105, and 0.0065 for AI Multimedia Lab. 

\subsection{Subtask 3: Privacy-preserving CT slice generation}

\subsubsection{Presented methods}

\noindent\textbf{csmorgan}~\cite{csmorgan} trained a compact DCGAN-style generator: a generator projecting $z \in \mathbb{R}^{128}$ into a $4\times4\times8c$ feature map ($c=48$) followed by six nearest-neighbour upsample-plus-convolution blocks, with GroupNorm and LeakyReLU, reaching 256$\times$256 output. Key design choices included: strong zero-centred $R_1$ gradient regularisation ($\gamma=100$), a deliberately weaker discriminator (half the base channel width, $10\times$ lower learning rate), exponential moving average (EMA) of generator weights ($\rho=0.999$), and patience-based early stopping (patience 30, minimum 80 epochs). The design prioritised empirical privacy over visual fidelity.
\\

\noindent\textbf{DS@GT ARC}~\cite{eric} combined a Differentially Private Optimal Transport Conditional Flow Matching (DP-OT-CFM) generator with a post-generation geometric filtering pipeline referred to as the ``Supervisor''. The core generator is a 34.5M-parameter UNet trained using OT-CFM, which regresses a time-dependent vector field between a Gaussian base distribution and the empirical CT data distribution via minibatch optimal transport coupling. For submissions involving differential privacy, the generator was trained with DP-SGD via the Opacus library (clipping norm 1.0, noise multiplier 1.1), corresponding to an estimated privacy budget of $(\varepsilon=5.64, \delta=10^{-5})$-DP. A candidate pool of 20,000 synthetic slices was generated at inference time before routing to the Supervisor for filtering down to the required 5,000 images.
The Supervisor scored each candidate by its distance from the training manifold in one of three learned latent spaces: a \textit{spatial autoencoder} using cosine distance; a \textit{contrastive autoencoder} using geodesic distance over a $k$-NN graph on PCA-whitened descriptors; and a \textit{Riemannian autoencoder} using an anisotropic metric decomposing displacement into tangent and normal components. Final subset selection applied either Determinantal Point Processes (DPP) or Stein Kernel Thinning (SKT) to enforce diversity. Ten configurations were submitted in total, systematically combining the presence or absence of DP-SGD training, the choice of Supervisor encoder (spatial, geodesic, or Riemannian), and the choice of coreset selection method (DPP, SKT, weighted DPP, or greedy), alongside a non-private baseline without any Supervisor filtering (\texttt{fm-unet-base}).

Ten configurations were submitted, combining the presence or absence of DP-SGD training, the choice of Supervisor encoder, and the choice of coreset selection method. The baseline run \texttt{fm-unet-base} used the non-private OT-CFM generator without any Supervisor filtering. Runs \texttt{sv-spatial-dpp} and \texttt{sv-spatial-kt} applied the spatial autoencoder Supervisor with DPP and kernel thinning selection respectively, without DP training. Runs \texttt{sv-geodesic-wdpp} and \texttt{sv-geodesic-skt} applied the geodesic (contrastive) Supervisor with weighted DPP and SKT selection respectively, also without DP training. Run \texttt{fm-unet-dp} used DP-SGD training alone, without any post-generation filtering. Runs \texttt{dp-sv-spatial-dpp} and \texttt{dp-sv-geodesic-greedy} combined DP-SGD training with spatial and geodesic Supervisor filtering respectively, the latter using greedy distance-based selection. Finally, runs \texttt{dp-riemannian-wdpp} and \texttt{dp-riemannian-skt} combined DP-SGD training with the Riemannian autoencoder Supervisor and weighted DPP or SKT selection respectively.
\\

\noindent\textbf{AI Multimedia Lab}~\cite{AIM} trained StyleGAN2 on the provided development set, using non-saturating logistic loss with path length regularisation for the generator and R1 gradient penalty ($\gamma=10$) for the discriminator. Training followed a progressive growing schedule from $8\times8$ to $256\times256$ resolution.

\subsubsection{Results}

Table~\ref{tab:subtask3} reports official results for Subtask 3. Per-attack leakage scores and composite PPS are shown for all submitted runs with working notes.

\begin{table}[ht]
\caption{Results of participant submissions for Subtask 3: Privacy-Preserving CT Slice Generation. (* task organising team)}
\label{tab:subtask3}
\centering
\begin{tabular}{llrrrrrr}
\toprule
\# & Participant / Run & FID $\downarrow$ & PPS $\uparrow$ & $\ell_1$ (NND) & $\ell_2$ (MIA) & $\ell_3$ (Re-ID) & $\ell_4$ (Inv.) \\
\midrule
1  & csmorgan                                  & 3.2358 & 0.9086 & 0.0213 & 0.1832 & 0.1611 & 0.0000 \\
2  & DS@GT ARC -- fm-unet-dp                   & 0.3290 & 0.5490 & 0.0901 & 0.3797 & 0.9933 & 0.3410 \\
3  & DS@GT ARC -- dp-sv-spatial-dpp            & 0.2639 & 0.4920 & 0.0835 & 0.3519 & 0.9732 & 0.6250 \\
4  & DS@GT ARC -- dp-sv-geodesic-greedy        & 0.4272 & 0.4980 & 0.0804 & 0.3725 & 0.9866 & 0.5620 \\
5  & DS@GT ARC -- dp-riemannian-wdpp           & 0.3250 & 0.4660 & 0.0880 & 0.3838 & 0.9933 & 0.6710 \\
6  & DS@GT ARC -- dp-riemannian-skt            & 0.3758 & 0.4330 & 0.0874 & 0.3724 & 0.9933 & 0.8140 \\
7  & DS@GT ARC -- sv-geodesic-wdpp             & 0.6355 & 0.4500 & 0.1161 & 0.5180 & 1.0000 & 0.5650 \\
8  & DS@GT ARC -- sv-spatial-kt                & 0.5645 & 0.4490 & 0.1103 & 0.4775 & 0.9933 & 0.6270 \\
9  & DS@GT ARC -- sv-spatial-dpp               & 0.5495 & 0.3800 & 0.1124 & 0.4902 & 0.9933 & 0.8870 \\
10 & DS@GT ARC -- sv-geodesic-skt              & 0.9542 & 0.3800 & 0.1040 & 0.4678 & 1.0000 & 0.9070 \\
11 & DS@GT ARC -- fm-unet-base                 & 0.3385 & 0.3580 & 0.1459 & 0.5916 & 0.9933 & 0.8370 \\
*  & AI Multimedia Lab                         & 1.2082 & 0.6041 & 0.1283 & 0.3514 & 0.9060 & 0.1964 \\
\bottomrule
\end{tabular}
\end{table}

\begin{figure}[!tp]
     \centering
     \begin{subfigure}[b]{\textwidth}
         \centering
         \includegraphics[width=14cm]{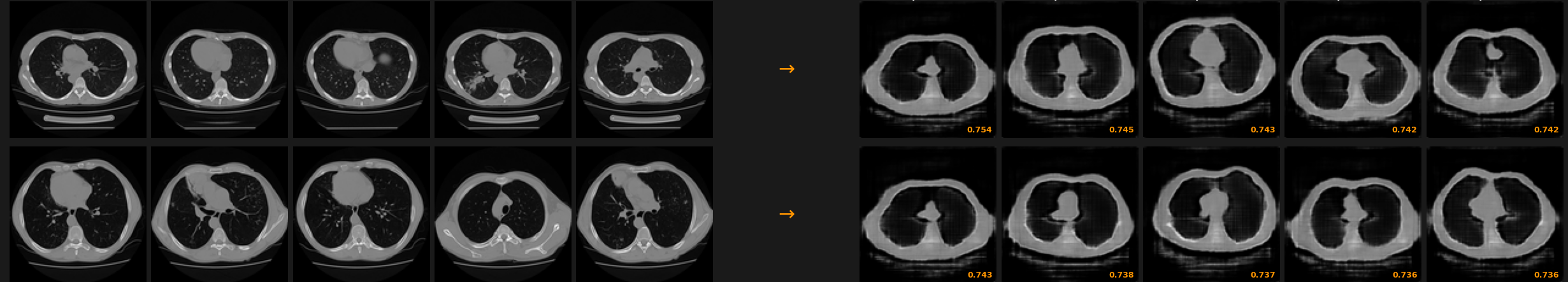}
         \caption{Real training slices and their closest synthetic matches per patient (csmorgan).}
         \label{fig:ex1-csmorgan}
     \end{subfigure}
     \\
          \begin{subfigure}[b]{\textwidth}
         \centering
         \includegraphics[width=14cm]{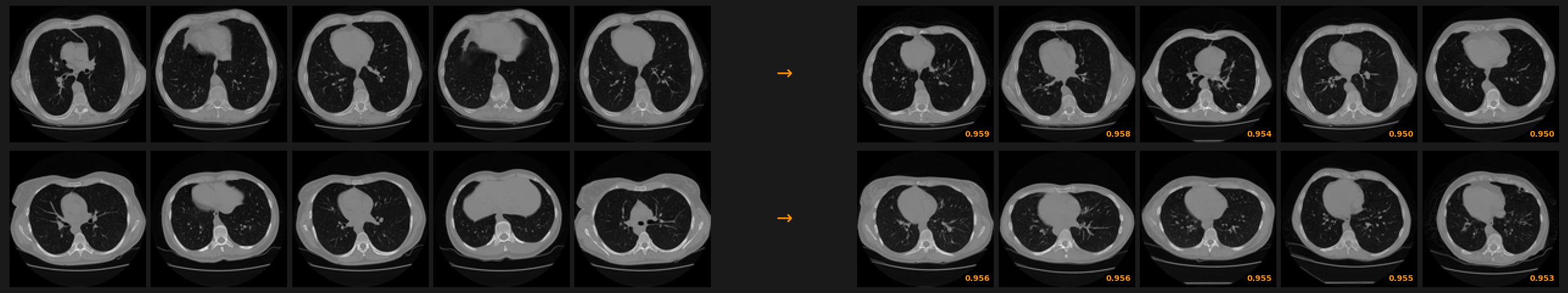}
         \caption{Real training slices and their closest synthetic matches per patient (DS@GT ARC - dp-sv-spatial-dpp).}
         \label{fig:sub-dp-sv-spatial-dpp}
     \end{subfigure}
     \\
          \begin{subfigure}[b]{\textwidth}
         \centering
         \includegraphics[width=14cm]{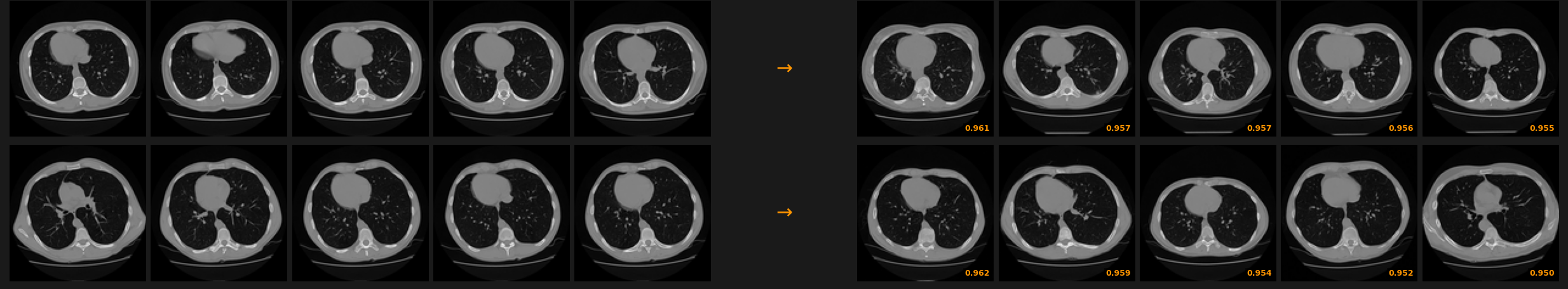}
        \caption{Real training slices and their closest synthetic matches per patient (DS@GT ARC - dp-riemannian-skt).}
         \label{fig:sub-dp-riemannian-skt}
     \end{subfigure}
     \\
          \begin{subfigure}[b]{\textwidth}
         \centering
         \includegraphics[width=14cm]{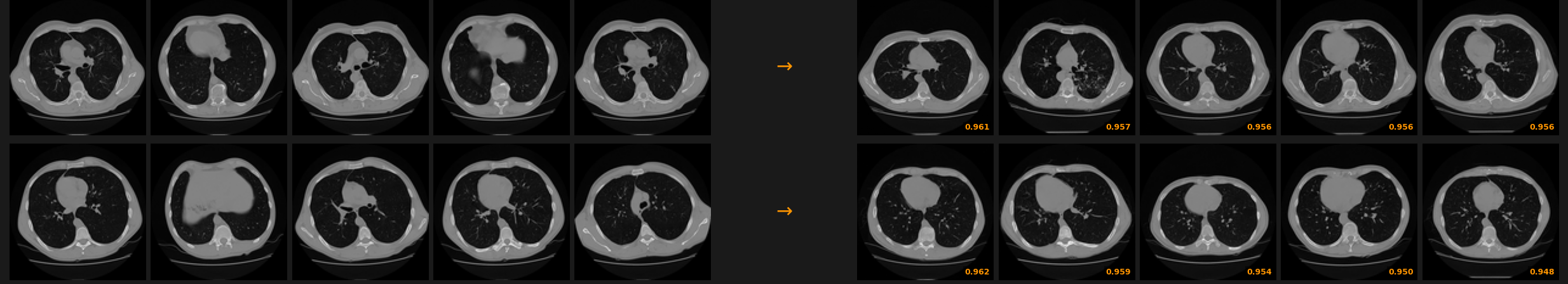}
        \caption{Real training slices and their closest synthetic matches per patient (DS@GT ARC - dp-riemannian-wdpp).}
         \label{fig:sub-dp-riemannian-wdpp}
     \end{subfigure}
     \\
          \begin{subfigure}[b]{\textwidth}
         \centering
         \includegraphics[width=14cm]{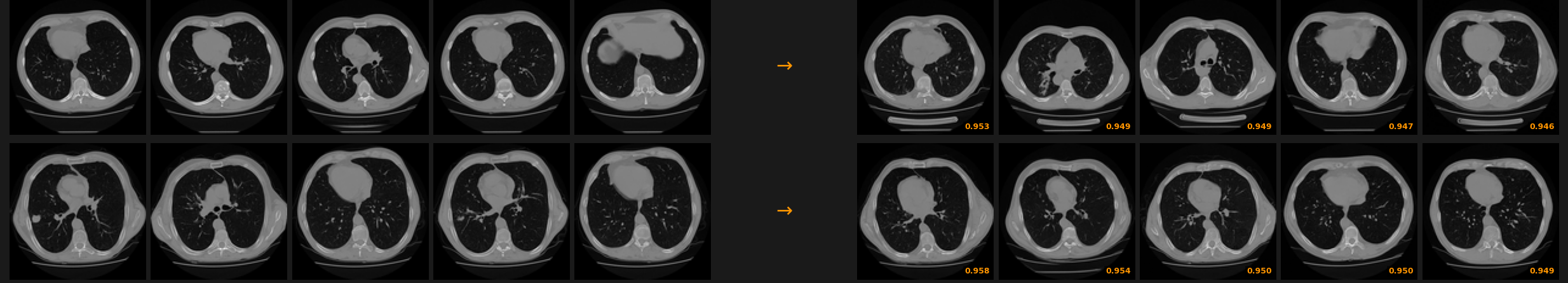}
         \caption{Real training slices and their closest synthetic matches per patient (DS@GT ARC - dp-sv-geodesic-greedy).}
         \label{fig:sub-dp-sv-geodesic-greedy}
     \end{subfigure}
     \\
  \begin{subfigure}[b]{\textwidth}
         \centering
         \includegraphics[width=14cm]{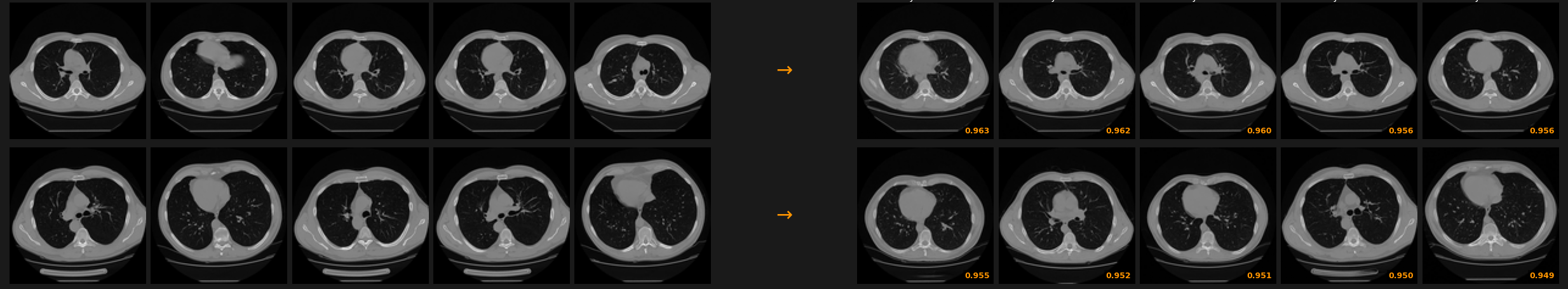}
         \caption{Real training slices and their closest synthetic matches per patient (DS@GT ARC - fm-unet-base).}
         \label{fig:sub-fm-unet-base}
     \end{subfigure}
     \\
      \begin{subfigure}[b]{\textwidth}
         \centering
         \includegraphics[width=14cm]{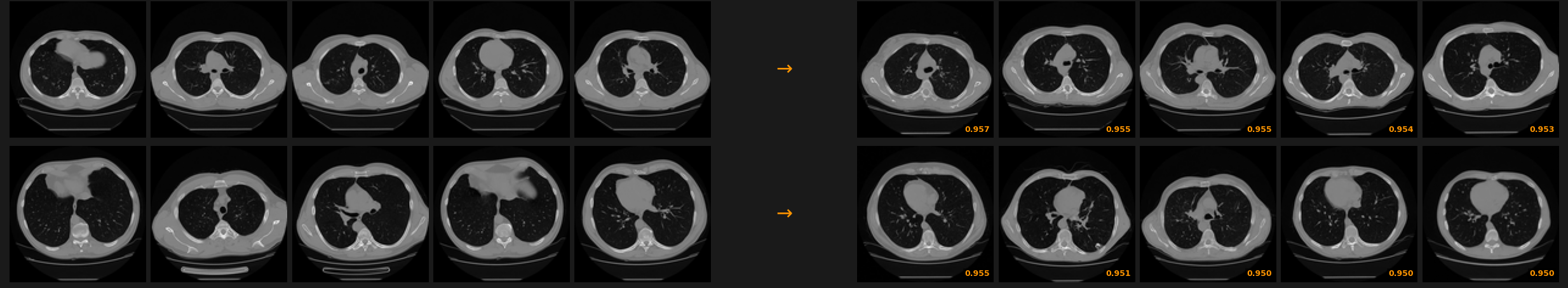}
         \caption{Real training slices and their closest synthetic matches per patient (DS@GT ARC - fm-unet-dp).}
         \label{fig:sub-fm-unet-dp}
     \end{subfigure}
     \\
         \caption{Real training slices alongside their most similar synthetic counterparts generated by each team's submission, as identified by the Patient Re-Identification attack (Attack 3) - part 1.}
        \label{fig:ex1}
\end{figure}

\begin{figure}[!tp]
     \centering    
        \begin{subfigure}[b]{\textwidth}
         \centering
         \includegraphics[width=14cm]{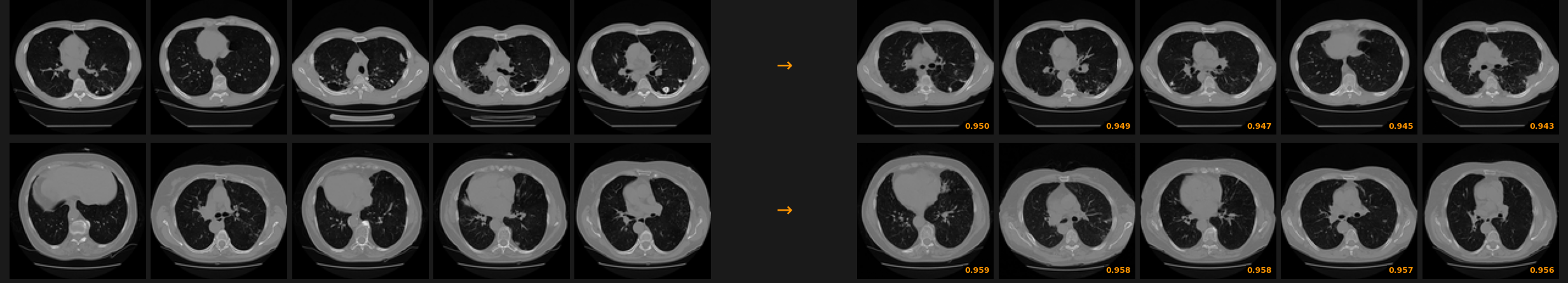}
         \caption{Real training slices and their closest synthetic matches per patient (DS@GT ARC - sv-geodesic-skt).}
         \label{fig:sub-sv-geodesic-skt}
     \end{subfigure}
     \\
          \begin{subfigure}[b]{\textwidth}
         \centering
         \includegraphics[width=14cm]{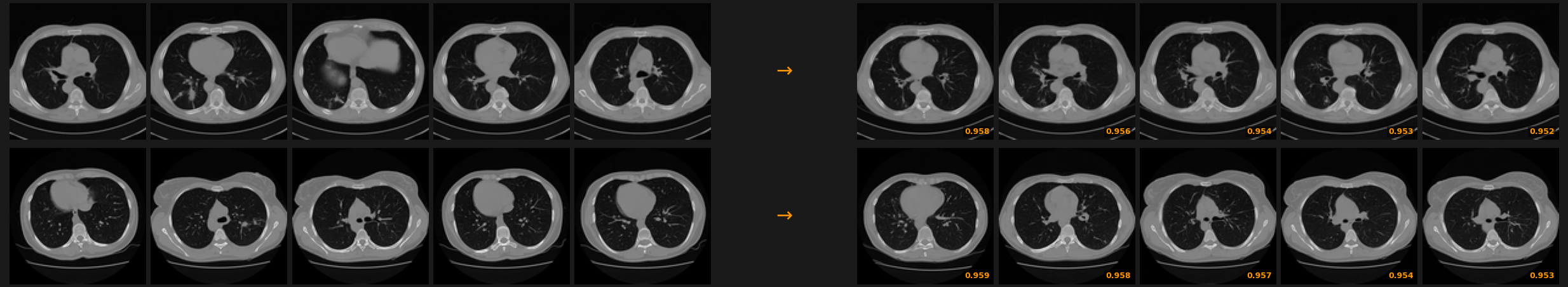}
         \caption{Real training slices and their closest synthetic matches per patient (DS@GT ARC - sv-geodesic-wdpp).}
         \label{fig:sub-sv-geodesic-wdpp}
     \end{subfigure}
     \\
               \begin{subfigure}[b]{\textwidth}
         \centering
         \includegraphics[width=14cm]{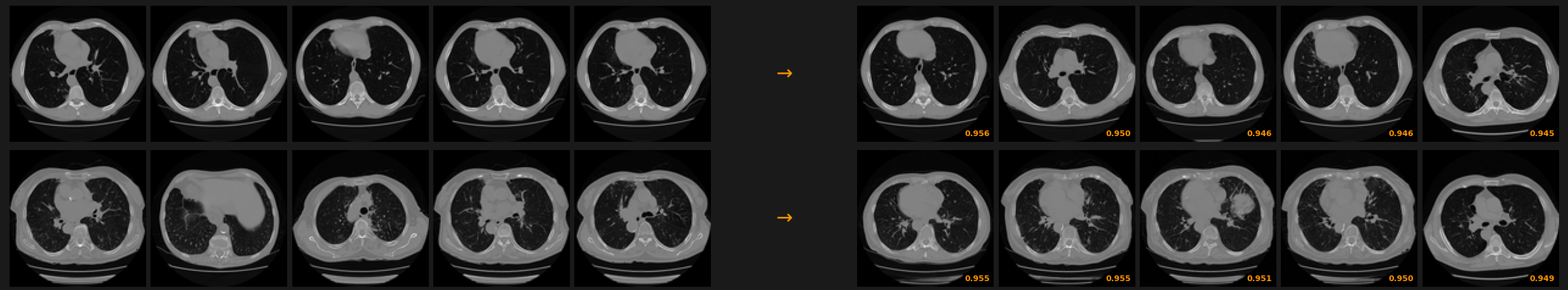}
          \caption{Real training slices and their closest synthetic matches per patient (DS@GT ARC - sv-spatial-dpp).}
         \label{fig:sub-sv-spatial-dpp}
     \end{subfigure}
     \\
               \begin{subfigure}[b]{\textwidth}
         \centering
         \includegraphics[width=14cm]{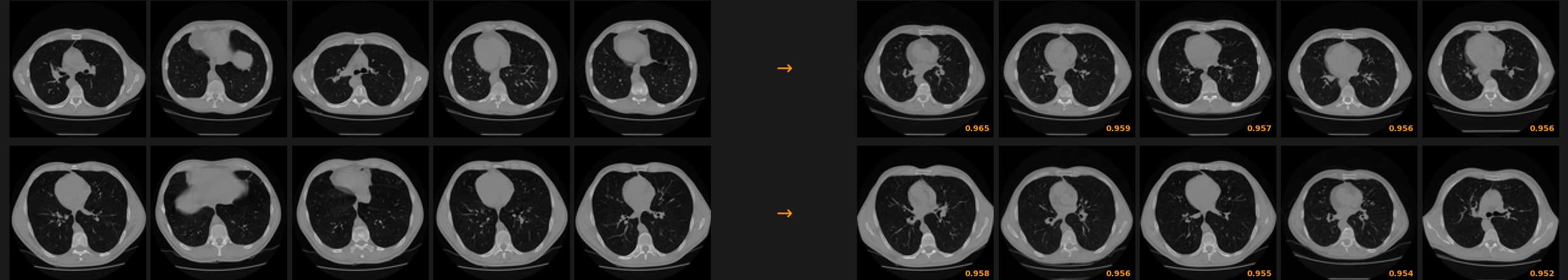}
         \caption{Real training slices and their closest synthetic matches per patient (DS@GT ARC - sv-spatial-kt).}
         \label{fig:sub-sv-spatial-kt}
     \end{subfigure}
          \\

     \begin{subfigure}[b]{\textwidth}
         \centering
         \includegraphics[width=14cm]{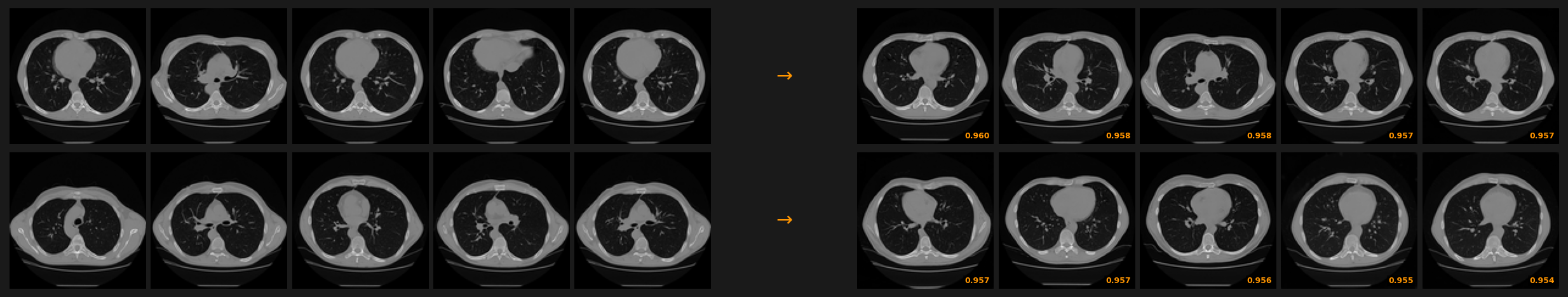}
         \caption{Real training slices and their closest synthetic matches per patient (AI Multimedia Lab).}
         \label{fig:sub-AIM}
     \end{subfigure}
     
         \caption{Real training slices alongside their most similar synthetic counterparts generated by each team's submission, as identified by the Patient Re-Identification attack (Attack 3) - part 2.}
        \label{fig:ex2}
\end{figure}

Figure~\ref{fig:ex1} and Figure~\ref{fig:ex2} present, for each patient, real training slices alongside their most similar synthetic counterparts generated by each team's submission. The synthetic images were selected and ranked by cosine similarity to the per-patient anatomical centroid, as computed by the Patient Re-Identification attack (Attack 3), with the cosine similarity score reported for each synthetic image. Beyond re-identification, these examples also allow a qualitative assessment of the visual and anatomical fidelity of the generated images.

Examining Figure~\ref{fig:ex1} together with Table~\ref{tab:subtask3}, the csmorgan submission achieved strong privacy scores across all four attacks, yielding the highest composite PPS of 0.9086 in this edition. However, this came at a clear cost in image quality: their submission obtained a CT-specific FID of 3.2358 on our domain-specific evaluator, a model trained on real CT images from the same distribution as the training data. To contextualise this value, submissions producing visually and anatomically plausible synthetic CT slices achieved FID scores in the range of $0.26-0.64$ on the same evaluator. A FID of 3.2358 indicates substantial divergence from the real CT distribution, suggesting that the generated images do not faithfully reproduce the anatomical structures and visual characteristics of real lung CT slices, as is also visible upon inspection of Figure~\ref{fig:ex1}.a. Taken together, the high PPS and high FID of this submission illustrate the privacy-utility trade-off. However, the results in Table~\ref{tab:subtask3} also show that this trade-off is not absolute. The DS@GT ARC submission dp-sv-spatial-dpp achieved the best FID of 0.2639 while maintaining a PPS of 0.492, and fm-unet-dp obtained a competitive FID of 0.3290 with a PPS of 0.549 --- the highest among DS@GT ARC runs. The AI Multimedia Lab submission achieved a strong FID of 1.2082 with a PPS of 0.6041, placing it between the two extremes. Across all submissions, Attack 3 (Patient Re-Identification, $\ell_3$) emerges as the dominant and most persistent leakage vector: all submissions except csmorgan obtained $\ell_3$ values above 0.90, with several reaching 1.0000, indicating that high image fidelity is consistently accompanied by near-complete patient-level identity leakage regardless of the privacy mechanism employed.

Across their 10 submissions, DS@GT ARC explored the effect of combining DP-SGD training with post-generation geometric filtering on the privacy-utility trade-off. The non-private baseline \texttt{fm-unet-base} obtained a competitive FID of 0.3385 but the lowest PPS of 0.358, driven primarily by high leakage across Attacks 2 ($\ell_2=0.5916$), 3 ($\ell_3=0.9933$), and 4 ($\ell_4=0.837$). Introducing DP-SGD training substantially improved both the MIA and inversion attack scores: \texttt{fm-unet-dp} reduced $\ell_2$ to 0.3797 and $\ell_4$ to 0.341 while maintaining a similar FID of 0.3290, yielding the highest PPS among all DS@GT ARC runs (0.549). A notable finding is that DP-SGD also improved image quality relative to the baseline, with \texttt{dp-sv-spatial-dpp} achieving the best FID of 0.2639 across all submissions in this edition, which the team attributed to the gradient clipping acting as a regulariser on the large UNet architecture. Regarding Attack 1 (NND), all DP-trained runs flagged fewer memorised images than the baseline (489--679 vs 1,790 flagged images), with \texttt{dp-sv-geodesic-greedy} achieving the lowest $\ell_1$ of 0.0804 across all DS@GT ARC submissions.
However, Attack 3 (Patient Re-Identification) remained the dominant and most persistent leakage vector across all runs, with rank-1 patient coverage ranging from 0.9732 to 1.0000 and leakage scores between 0.9732 and 1.0000 regardless of the privacy mechanism employed. This indicates that neither DP-SGD training nor geometric filtering in the Supervisor pipeline was sufficient to prevent the generator from internalising the anatomical identity space of the 149 training patients. Runs without DP training but with Supervisor filtering (\texttt{sv-spatial-dpp}, \texttt{sv-spatial-kt}, \texttt{sv-geodesic-wdpp}, \texttt{sv-geodesic-skt}) achieved moderate reductions in $\ell_1$ and $\ell_2$ compared to the baseline, confirming that output-level filtering effectively reduces instance-level memorisation and distributional bias, but with $\ell_3$ remaining above 0.99 in all cases. Among the Riemannian autoencoder configurations, \texttt{dp-riemannian-wdpp} achieved a strong balance between FID (0.3250) and privacy (PPS=0.466), while \texttt{dp-riemannian-skt} suffered from higher inversion leakage ($\ell_4=0.814$), suggesting that the SKT selection strategy may inadvertently favour samples that are more easily invertible by the white-box attack.
\section{Discussion}
\label{sec:discussion}

\subsection{Subtask 1: Detect training data usage}

A wide range of strategies were deployed for Subtask 1, spanning transfer learning classifiers, handcrafted feature ensembles, autoencoder-based anomaly detection, and $k$-NN distance pipelines. Despite this variety, all submitted runs achieved Cohen's $\kappa$ scores close to or below zero, confirming that the task remains one of the most challenging in the series.

Bharathi \& Harish achieved development-phase performance of $\kappa=0.9122$ using a single ResNet34 model, but this did not transfer to the final test set, where both submitted runs yielded negative kappa values ($-0.0383$ and $-0.0776$). The team attributed this gap to domain shift and overfitting; however, domain shift is unlikely in this task setup, as the development and test real images are drawn from the same cohort of lung tuberculosis CT scans acquired under the same conditions. A more plausible explanation is overfitting: models trained to near-perfect accuracy on the development data had likely memorised distributional artefacts specific to that split rather than learning generalisable fingerprint features. Notably, the ensemble of three models performed worse than the single model, suggesting that all three architectures learned correlated, non-complementary representations under identical training conditions, so that soft voting did not provide the expected generalisation benefit.

The csmorgan team reported an internal validation AUC-ROC of 0.7557 using a multi-signal ensemble (spectral, statistical, deep $k$-NN, and shadow-autoencoder features), yet achieved $\kappa$ values between $-0.080$ and $-0.029$ on the official test set. This development-to-test gap indicates that internal cross-validation may have inflated performance estimates due to calibration drift or distributional differences between the development and hidden test sets.

The CVG-IBA team found that their ResNet-18 classifier achieved a cross-validated accuracy of 52.81\%, marginally above chance, while the pixel-statistics Random Forest and autoencoder ensemble both hovered at or below chance. When combined, ensemble configurations degraded rather than improved over the best individual method, suggesting that the weaker detectors introduced noise that diluted the ResNet-18 signal. The official submission (E3, Pixel-RF + AE-Ensemble) achieved $\kappa=-0.0243$, reflecting highly imbalanced specificity (87.2\%) versus sensitivity (10.4\%), consistent with a model defaulting to the ``not used'' class.

The AI Multimedia Lab's $k$-NN distance pipeline in ResNet-50 feature space yielded $\kappa=-0.0056$, essentially at chance, with perfectly balanced sensitivity and specificity around 50\%. The team attributed this to two factors: the lack of domain adaptation of the ResNet-50 features to CT data, and the possibility that the target GAN generalises sufficiently well to erase detectable instance-level signals from the generated distribution.

Cohen's $\kappa$ scores across all Subtask 1 submissions were consistently near zero or negative. This may also reflect the capability of the used GAN model to generate images that do not contain easily detectable ``fingerprints'' traceable to specific training samples, which represents a positive result from a privacy protection perspective.

\subsection{Subtask 2: Identify corresponding latent vectors}

Subtask 2 was new in this edition and produced contrasts in performance across methods. The key dividing line was between regression-based approaches and contrastive or unsupervised alternatives.

Agentili team achieved a perfect matching accuracy of $1.0000$ in their best submission. Their system demonstrated that direct image-to-latent regression with EfficientNetV2-S, combined with a cosine-dominant hybrid loss (Huber + cosine embedding), Optuna hyperparameter tuning, 5-fold cross-validation ensembling, and global Hungarian assignment, provides a highly effective and reliable solution. ViT-B/16 was found to be entirely ineffective for this task (Top-1 $\approx 0.75\%$), suggesting that convolutional inductive biases are essential for capturing the spatial regularities in synthetic CT images at this resolution. The reformulation of inference as a bipartite assignment problem was identified as the single most impactful methodological decision, dramatically outperforming greedy argmax matching.

The CVG-IBA team achieved 0.3677 with their E5 ensemble (three ResNet-50 regression objectives plus a dual-encoder contrastive model), confirming that combining complementary loss functions (MSE, cosine, combined) captures different aspects of latent geometry and their averaging before Hungarian matching reduces effective error rate. The standalone A4 contrastive encoder achieved only 2.12\% in cross-validation, consistent with the AI Multimedia Lab's finding, but contributed a complementary signal when added to the regression ensemble.

The csmorgan memory-bank InfoNCE matcher achieved a best result of 2.24\% (224 out of 10,000 correct matches), Training and validation curves showed clear overfitting to the development image-latent correspondences. The team identified possible generator-distribution shift between development and test data, non-injectivity of the generator, and optimistic evaluation caused by memory-bank reuse as contributing factors.

The AI Multimedia Lab's Method 1 (supervised ResNet-50 regression with combined MSE and cosine loss and Hungarian assignment) achieved 0.3737 matching accuracy, the second best result among non-perfect submissions. The four-method ablation confirmed that direct supervised regression consistently outperforms contrastive metric learning (Method 2, 0.0666) and training-free linear approximations in DINO feature space (Method 4, 0.0105). Hard negative mining with a randomly initialised memory bank (Method 3, 0.0065) proved counterproductive, suggesting that early training instability when negatives are uninformative outweighs the benefit of harder negatives later.

Overall, Subtask 2 results demonstrate that BigGAN-Deep's latent-to-image mapping is highly stable and predictable: a well-designed regression system can recover it with near-perfect accuracy. This raises important privacy implications, as near-perfect latent recoverability suggests that the generator does not adequately obscure the relationship between internal representations and generated outputs.

\subsection{Subtask 3: Privacy-preserving CT slice generation}
Subtask 3 revealed a fundamental tension between image fidelity and privacy preservation, with different teams occupying clearly distinct positions on the privacy--utility frontier. Results in Table~\ref{tab:subtask3} show that no submission achieved simultaneously strong privacy preservation and high image fidelity across all four attack surfaces, highlighting the difficulty of the task.

The csmorgan team's compact DCGAN-style generator achieved the strongest empirical privacy score (PPS = 0.9086), placing it at the privacy-extreme end of the spectrum. The near-zero white-box inversion leakage ($\ell_4 = 0.0000$) and low NND ($\ell_1 = 0.0213$) and MIA ($\ell_2 = 0.1832$) scores confirm that the combination of strong $R_1$ regularisation, weak discriminator, EMA inference, and early stopping effectively suppressed instance-level copying and latent-space memorisation. Crucially, and unlike all other submissions, patient re-identification leakage was also relatively low ($\ell_3 = 0.1611$), suggesting that the deliberate capacity constraints imposed on this generator prevented it from internalising the full anatomical identity space of the training cohort. However, this strong privacy came at a clear utility cost: the submission obtained a CT-specific FID of 3.2358, substantially higher than the 0.26--0.64 range achieved by other submissions, indicating that the generated images do not faithfully reproduce the anatomical and visual characteristics of real CT slices.

The DS@GT ARC team explored the most methodologically diverse range of configurations across ten submissions, combining DP-OT-CFM training with multiple geometric filtering strategies. Their best PPS of 0.549 was achieved by \texttt{fm-unet-dp}, a differentially private UNet generator without post-generation filtering. Counter-intuitively, applying DP-SGD improved rather than degraded image quality relative to the non-private baseline (FID = 0.3290 for \texttt{fm-unet-dp} vs. 0.3385 for \texttt{fm-unet-base}), an effect the team attributed to DP-SGD's gradient clipping acting as a regulariser on the 34.5M parameter model. The Supervisor pipeline proved most effective against instance-level memorisation and distributional bias: \texttt{dp-sv-geodesic-greedy} achieved the lowest Attack 1 leakage score of 0.0804 across all DS@GT ARC runs, and DP-trained runs consistently flagged fewer memorised images (489-679) compared to the non-private baseline (1,790). However, patient re-identification (Attack 3) remained near its maximum across all eleven configurations ($\ell_3$ = 0.9732--1.0000), confirming that neither DP-SGD training nor output-level geometric filtering is sufficient to prevent the generator from internalising the anatomical identity space of the 149 training patients. Among the Riemannian autoencoder configurations, \texttt{dp-riemannian-wdpp} achieved a favourable balance between image quality (FID = 0.3250) and privacy (PPS = 0.466), while non-DP Supervisor runs without DP training (\texttt{sv-spatial-dpp}, \texttt{sv-geodesic-skt}) showed the weakest overall privacy scores (PPS = 0.380), driven by high leakage across Attacks 2, 3, and 4 simultaneously.

The AI Multimedia Lab's StyleGAN2 submission achieved a FID of 1.2082 with a moderate PPS of 0.6041. Attack analysis revealed a nuanced leakage profile: instance-level copying was limited ($\ell_1 = 0.1283$, NND ratio = 0.8717, with 1,359 flagged images), and latent-space memorisation was also relatively low ($\ell_4 = 0.1964$, reconstruction delta = 0.9085). The MIA leakage was moderate ($\ell_2 = 0.3514$, AUC-ROC = 0.6757, TPR@5\%FPR = 22.5\%), indicating that a distance-based adversary can achieve meaningful but not severe membership separation. However, patient re-identification was near-complete ($\ell_3 = 0.9060$), with 90.6\% of the 149 training patients appearing as the top cosine-similarity match for at least one synthetic image, 135$\times$ above the chance level of 0.0067. This pattern consisting in limited slice-level copying and modest latent-space memorisation alongside near-complete patient identity exposure illustrates a critical gap between incidental privacy and formal privacy guarantees: the StyleGAN2 generator internalised the full identity space of training patients without reproducing individual slices.

A particular observation concerns Attack 3 (Patient Re-Identification), whose leakage scores were consistently near-maximal across all submissions that produced anatomically plausible images. While this attack measures a genuine linkage risk, the results raise the question of whether high rank-1 patient coverage is inherently a privacy failure or an expected consequence of a well-generalising generative model. A generator trained on data from a finite cohort of patients and asked to produce a large synthetic set will naturally span the full anatomical variation of its training distribution; by design, this means every patient's anatomy will be represented somewhere in the synthetic output. This interpretation is supported by the consistently high normalised assignment entropy (0.91--0.95 across all submissions), indicating broad and uniform coverage of patient identities rather than targeted memorisation of specific individuals. Notably, the only submission achieving low $\ell_3$ (csmorgan, $\ell_3 = 0.1611$) did so at the cost of a substantially degraded FID of 3.2358, suggesting that low re-identification leakage was a consequence of failing to learn the real data distribution rather than of effective privacy protection. These observations suggest that Attack 3, in its current formulation, may conflate distributional coverage with identity leakage in settings with small training cohorts, and that its interpretation should be considered alongside image quality metrics and the other three attack scores rather than in isolation.
The behaviour and validity of this attack as a privacy metric will be studied more carefully in future editions of the task, with the goal of better 
distinguishing between genuine patient-level memorisation and the expected distributional coverage of a well-trained generative model.
\section{Lessons learned}
\label{sec:lessons}

The 2026 edition of the ImageCLEFmedical GANs task provided valuable insights into the challenges of detecting training data exposure in synthetic medical images, with each subtask offering distinct lessons from a different angle on the privacy leakage problem.

\textbf{Subtask 1} continued to demonstrate the fundamental hardness of image-level membership inference against well-regularised GANs under black-box conditions. Despite the use of a wide array of methods, the highest Cohen's $\kappa$ scores remained close to or below zero. This may simultaneously indicate that current detection frameworks are insufficiently sensitive and that the target GAN has effectively generalised beyond per-sample memorisation. A key lesson is that development-phase performance -- whether cross-validated accuracy or internal AUC-ROC -- is an unreliable predictor of challenge-set generalisation for this task, as multiple teams reported dramatic performance drops between development and test phases.

\textbf{Subtask 2} introduced a new and important finding: the BigGAN-Deep latent-to-image mapping is highly stable and can be recovered with perfect accuracy using appropriate regression-based methods. The right combination of backbone architecture, loss function design (cosine-dominant hybrid objective), global bipartite assignment (Hungarian matching), and hyperparameter optimisation is sufficient to achieve perfect or near-perfect latent recovery. This has direct privacy implications: if an adversary can access generated images and knows the generator was BigGAN-Deep, they can recover the latent codes with high accuracy, potentially enabling further reconstruction or attribution attacks. The failure of contrastive and unsupervised methods in this subtask also provides practical guidance for future participants.

\textbf{Subtask 3} exposed privacy-utility trade-off in synthetic medical imaging. Nearly all teams achieved low instance-level memorisation scores, confirming that direct image copying is avoidable with standard regularisation. However, patient re-identification remained the dominant leakage vector across all configurations, suggesting that generative models trained on real patient cohorts internalise the full distribution of patient anatomies even without reproducing individual slices. Formal privacy mechanisms such as DP-SGD provide mathematically bounded protection and, in this edition, were found to improve rather than degrade image quality for our specific model scale, though the re-identification vulnerability persisted.

Collectively, the three subtasks show that generative model outputs can leak different types of information depending on the model architecture and the nature of the detection task. Membership inference (Subtask 1) probes instance-level leakage and remains largely undetectable for well-generalised GANs. Latent matching (Subtask 2) reveals structural predictability of the learned mapping and is highly solvable with supervised regression. Privacy-preserving generation (Subtask 3) highlights the challenge of eliminating patient-level identity leakage at the distributional level. Continued benchmarking and task diversification will be essential for a more comprehensive understanding of privacy risks in generative medical imaging.

\section{Conclusions}
\label{sec:conclusions}

The 2026 edition of the ImageCLEFmedical GANs Task further investigated the privacy and security implications of GAN-generated synthetic medical images by addressing three complementary challenges: detecting training data membership (Subtask 1), recovering latent vectors from generated images (Subtask 2), and generating privacy-preserving CT slices (Subtask 3).

In Subtask 1, results were consistent with previous editions: despite a broad range of approaches -- from transfer learning classifiers and handcrafted feature ensembles to $k$-NN distance pipelines and multi-signal feature ensembles -- all methods performed at or below chance level on the official test set, with Cohen's $\kappa$ values ranging from -0.0804 to -0.0056. A recurring finding was the large gap between development-phase and test-phase performance, observed across multiple teams, indicating that current approaches tend to overfit to development-set artefacts rather than learning generalisable membership signals. These results may simultaneously reflect the limitations of existing detection frameworks and the capability of the target GAN to produce images that do not retain easily detectable training-data fingerprints, which would represent an encouraging outcome from a privacy standpoint.

In Subtask 2, the new latent-matching task produced a range of outcomes, spanning from near-random performance to perfect accuracy. The Agentili team demonstrated that supervised image-to-latent regression with EfficientNetV2-S, a cosine-dominant hybrid loss, Optuna hyperparameter optimisation, 5-fold cross-validation ensembling, and Hungarian matching achieves a perfect matching accuracy of 1.0000 on the 10,000-way problem, correctly identifying all image-latent correspondences. This confirms that BigGAN-Deep's latent-to-image mapping is highly stable and predictable, a result with direct privacy implications: an adversary with access to generated images can recover the originating latent codes with high accuracy, potentially enabling further reconstruction or attribution attacks. Regression-based approaches consistently and substantially outperformed contrastive and unsupervised alternatives across all participating teams.

In Subtask 3, results revealed a clear privacy--utility trade-off across submissions, with no team achieving simultaneously strong privacy preservation and high image fidelity across all four attack surfaces. Patient re-identification (Attack 3) emerged as the most persistent leakage vector, with all high-fidelity submissions scoring above 0.90 regardless of the privacy mechanism employed, whether DP-SGD, geometric filtering, or neither. As discussed, however, the validity of Attack 3 as a standalone privacy metric in settings with small training cohorts warrants further investigation, as high rank-1 patient coverage may partly reflect the expected distributional coverage of a well-trained generative model rather than targeted memorisation. This will be studied in future editions of the task.

Overall, the 2026 edition underscores the need for ongoing research into privacy-preserving generative modelling, particularly in sensitive domains such as healthcare. The three subtasks collectively illustrate that privacy leakage in synthetic medical images is a multi-faceted problem: membership signals may be undetectable at the image level while latent representations remain fully recoverable, and instance-level copying can be suppressed without eliminating patient-level identity exposure. Benchmarking challenges like this remain crucial for building a more comprehensive and nuanced understanding of the privacy risks associated with synthetic medical data. We will continue investigating these issues in future editions of ImageCLEF, with plans to refine the evaluation protocol and introduce new tasks aimed at further exploring the boundaries of privacy, utility, and generalisation in generative medical imaging.

\section*{Declaration on Generative AI}

During the preparation of this work, the authors used Claude (Anthropic) to assist with drafting and polishing specific sections of the manuscript. All scientific content and references were provided by the authors, who reviewed and verified all AI-assisted text and take full responsibility for the content of the article.

\begin{acknowledgments}
The work of Alexandra-Georgiana Andrei and Dan-Cristian Stanciu was supported by a grant of the Ministry of Research, Innovation and Digitization, CCCDI -- UEFISCDI, project number PN-IV-P6-6.3-SOL-2024-3-0320.
\end{acknowledgments}

\bibliography{sample-ceur}

\end{document}